\documentclass{article}

\usepackage[english]{babel}
\usepackage{multirow}
\usepackage{graphicx}
\usepackage{subcaption}
\usepackage[utf8]{inputenc}% Set page size and margins
\usepackage[letterpaper,top=2cm,bottom=2cm,left=3cm,right=3cm,marginparwidth=1.75cm]{geometry}

\usepackage{amsmath}
\usepackage{colortbl}
\usepackage{xcolor}
\usepackage{pifont}
\usepackage{authblk}
\usepackage{amsfonts}
\usepackage{booktabs}
\usepackage{caption}
\usepackage{adjustbox}
\makeatletter
\newcommand{\printfnsymbol}[1]{%
  \textsuperscript{\@fnsymbol{#1}}%
}
\makeatother
\usepackage[colorlinks=true, allcolors=blue]{hyperref}
\newlength\savewidth\newcommand\shline{\noalign{\global\savewidth\arrayrulewidth\global\arrayrulewidth1pt}\hline\noalign{\global\arrayrulewidth\savewidth}}
\definecolor{baselinecolor}{gray}{.9}
\definecolor{darkgray}{gray}{.7}
\newcommand{\cb}{\cellcolor{green!30}}

\title{{\bf Chem42}\footnote{Chem42 is part of the Omics42 platform which includes also a family of genomic LMs and a family of proteins LMs named Gene42 and Prot42, respectively. Refer to \href{https://huggingface.co/spaces/inceptionai/Omics42}{\textbf{Omics42}} blog at \href{https://huggingface.co/inceptionai}{huggingface.co/inceptionai} for further details.} : a Family of chemical Language Models for Target-aware Ligand Generation} 

\author[1]{Aahan Singh\thanks{These authors contributed equally to this work.}}
\author[2]{Engin Tekin\printfnsymbol{2}}
\author[1]{Maryam Nadeem\printfnsymbol{2}}
\author[1]{Nancy A. ElNaker}
\author[1]{Mohammad Amaan Sayeed}
\author[2]{Natalia Vassilieva}
\author[1]{Boulbaba {Ben Amor}\thanks{Corresponding author: Boulbaba Ben Amor \url{boulbaba.amor@inceptionai.ai}}}

\affil[1]{Inception Institute of Artificial Intelligence, Abu Dhabi, UAE.}
\affil[2]{Cerebras Systems, Sunnyvale, CA, USA.}

\date{}

\begin{document}
\maketitle

\begin{abstract}

Revolutionizing drug discovery demands more than just understanding molecular interactions—it requires generative models that can design novel ligands tailored to specific biological targets. While chemical Language Models (cLMs) have made strides in learning molecular properties, most fail to incorporate target-specific insights, restricting their ability to drive \textit{de-novo} ligand generation. Chem42, a cutting-edge family of generative chemical Language Models, is designed to bridge this gap. By integrating atomic-level interactions with multimodal inputs from Prot42, a complementary protein Language Model \cite{Prot42}, Chem42 achieves a sophisticated cross-modal representation of molecular structures, interactions, and binding patterns. This innovative framework enables the creation of structurally valid, synthetically accessible ligands with enhanced target specificity. Evaluations across diverse protein targets confirm that Chem42 surpasses existing approaches in chemical validity, target-aware design, and predicted binding affinity. By reducing the search space of viable drug candidates, Chem42 could accelerate the drug discovery pipeline, offering a powerful generative AI tool for precision medicine. The Chem42 models, available at \href{https://huggingface.co/inceptionai}{huggingface.co/inceptionai}, set a new benchmark in molecule property prediction, conditional molecule generation, and target-aware ligand design.

\begin{figure}[ht!] \centering \includegraphics[width=1\linewidth]{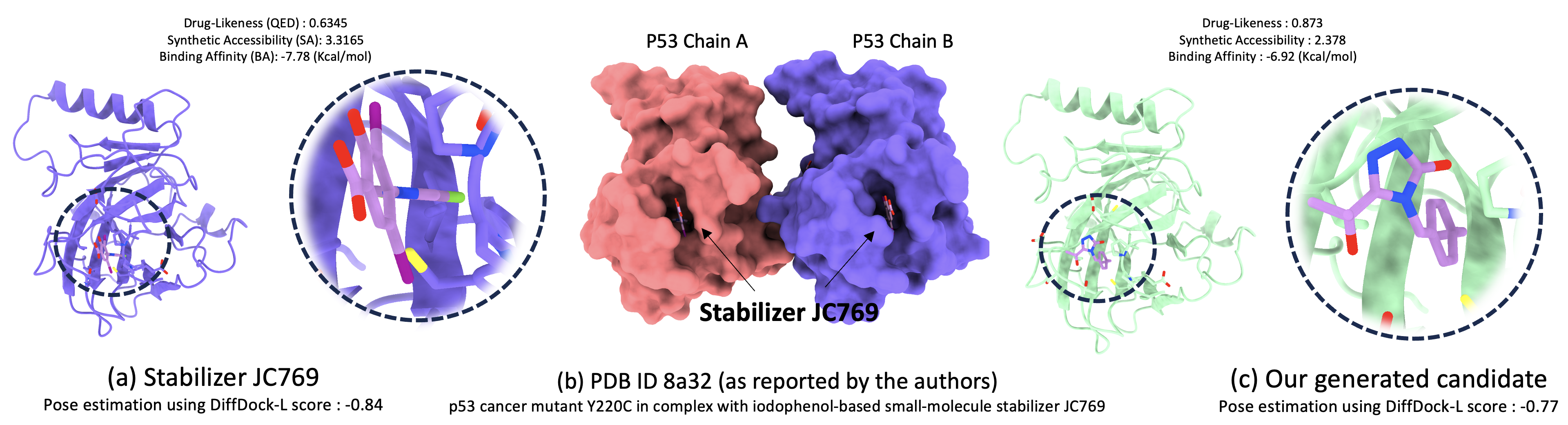} \caption{Example of a Chem42-generated ligand designed to bind the Y220C mutant p53 protein (PDB ID: 8a32), shown on the right (panel (c)). For comparison, panel (a) presents the binding pose of the recently proposed small-molecule stabilizer JC769 as estimated using DiffDock-L \cite{corso2024deep}, while panel (b) displays the stabilizer JC769 as originally reported in the literature \cite{stephenson2022discovery}. The protein target context is captured using our protein Language Model (Prot42 \cite{Prot42}) and integrated into the chemical Language Model (Chem42) to guide ligand generation. Calculated metrics—including Drug-likeness, Synthetic Accessibility, and Binding Affinity—demonstrate the effectiveness of Chem42 in achieving state-of-the-art molecular design results.} \label{fig:p53} \end{figure}

\end{abstract}

%\newpage
%\tableofcontents

\section{Introduction}
\label{sec:Introduction}

The discovery and design of novel therapeutic molecules remains one of the most critical challenges in drug development, requiring a precise understanding of molecular interactions and target specificity. Traditional computational methods, such as molecular docking and quantitative structure-activity relationship (QSAR) models, often struggle with the vast chemical space and the dynamic nature of molecular interactions \cite{sadybekov2023computational}. Recent advancements in Large Language Models (LLMs) have driven the emergence of chemical Language Models (cLMs) \cite{cai2024foundation, bagal2022molgpt, ahmad2022chemberta2chemicalfoundationmodels, Irwin_2022}, which learn molecular representations to generate novel structures and predict their properties. However, these models typically lack target-specific awareness, limiting their utility for rational ligand design.

Drug discovery is inherently \textbf{multimodal}, since ligand activity is dictated not only by the structure of a molecule, but also by its complex interactions with protein targets. To address this limitation, we introduce Chem42, a generative chemical-protein modeling framework that integrates our protein Language Model (pLM), named Prot42 \cite{Prot42}, with our chemical Language Model for target-aware ligand generation. This cross-modal approach enhances molecular design by embedding the protein context into the generative process, enabling the model to generate valid, biologically relevant, and target-specific ligands. Figure \ref{fig:p53} illustrates how Chem42 generates a ligand for the Y220C p53 mutant protein (a tumor suppressor called also "Guardian of the Genome"), comparing its binding pose to JC769, a known stabilizer (PDB ID: 8a32) \cite{stephenson2022discovery}. Both ligands bind to the same pocket, demonstrating the model’s capacity to design structure-guided molecules. Given a target protein such as p53, we prompt Prot42 with its sequence, embedding its molecular features into Chem42 via cross-attention, allowing for ligand generation guided by target-specific interactions. \textbf{To the best of our knowledge, this is the first framework where a cLM and a pLM collaborate to explicitly account for protein context in ligand generation, marking a significant advancement in AI-driven drug design.}

\section{Related Works}
\label{sec:SoTA}

This section reviews recent advances in chemical Language Models (cLMs) for molecule generation and property prediction, emphasizing the significance of tokenization strategies in developing effective pre-trained models. In addition, we discuss how these models facilitate advancements in drug discovery, particularly through approaches that enable target-aware ligand generation.  

\subsection*{Chemical Language Models for Molecular Generation}

Recent advances in chemical Language Models (cLMs) have significantly impacted molecular generation, leveraging deep learning to propose novel compounds with targeted properties. ChemFM \cite{cai2024foundation}, a large-scale foundation model built on the Tiny LLaMA architecture, employs self-supervised causal language modeling on a dataset of 178 million molecules, capturing complex molecular representations without task-specific architectures. With up to 3 billion parameters, ChemFM captures complex molecular representations and generalizes across multiple chemical tasks without relying on task-specific architectures. MolGPT \cite{bagal2022molgpt} adopts a GPT-inspired transformer-decoder framework utilizing masked self-attention on SMILES representations, emphasizing molecular generation tasks. Beyond ChemFM and MolGPT, various chemical LLMs have adopted different tokenization strategies and pre-training methodologies to enhance molecular design. SMILES-BERT \cite{wang2019smilesbert} and ChemBERTa \cite{Chithrananda2020ChemBERTaLS} employ masked language modeling (MLM) to refine molecular property prediction, while models such as Chemformer \cite{Irwin_2022} integrate MLM paired with an autoregressive token generation (ATG tokenization) strategy to improve both property prediction and molecule generation. MolXPT \cite{liu2023molxpt} expands the capabilities of chemical LLMs by incorporating textual descriptions of molecules, excelling in molecule captioning and molecular property prediction. DrugGPT \cite{li2023druggpt} introduces an autoregressive model based on GPT for the discovery of protein specific ligands.

Chemical Language Models (cLMs) employ distinct pretraining strategies depending on their primary objective—molecular property prediction or molecule generation. The two dominant approaches are Masked Language Modeling (MLM) and Autoregressive Token Generation (ATG), each optimized for different tasks. MLM, inspired by BERT-based models, trains by masking random tokens in molecular sequences and predicting them based on surrounding context. This bidirectional training approach enhances deep molecular understanding and improves property prediction. Notable examples include SMILES-BERT \cite{wang2019smilesbert} and ChemBERTa \cite{Chithrananda2020ChemBERTaLS}, both designed for molecular property prediction, while MG-BERT \cite{zhang2021mgbert} extends this method by integrating graph adjacency knowledge to refine molecular representations. ATG, on the other hand, generates molecular sequences token-by-token, predicting each token based on its predecessors. This method is particularly effective for de novo molecular design and reaction prediction. ChemGPT \cite{Frey2023} specializes in molecular sequence completion using ATG, while MolXPT \cite{liu2023molxpt} enhances generative tasks by combining SMILES representations with textual descriptions, excelling in molecular captioning and property prediction. Meanwhile, DrugGPT \cite{li2023druggpt}, built on GPT-2, focuses on controlled generation of novel drug-like molecules. Overall, MLM-based models excel in molecular property prediction, while ATG-based models are more effective for generative tasks \cite{liao2024wordsmoleculessurveylarge}. Chemformer \cite{Irwin_2022}, leveraging a fusion of both strategies, stands out as a powerful model for both molecule generation and molecular property prediction.

These models mark a paradigm shift from traditional rule-based approaches to scalable, data-driven strategies, enabling more efficient molecular generation, optimization, and property prediction. By integrating diverse tokenization techniques, pretraining objectives, and model architectures, AI-driven chemical design continues to evolve, pushing the boundaries of computational drug discovery and accelerating the identification of novel molecules with optimized properties.

% These models represent a shift from traditional rule-based approaches to scalable, data-driven strategies, enabling more efficient molecular generation, optimization, and property prediction. The combination of diverse tokenization techniques, pretraining objectives, and model architectures continues to push the boundaries of AI-driven chemical design, improving the efficiency and effectiveness of molecular discovery.

%\textcolor{red}{Elaborate here on the SMILES Language and Tokenization for each of these models - link to Section 3.3. - Refer to this paper (\url{https://arxiv.org/pdf/2402.01439}) for the 3 approaches.}

%\textcolor{red}{Several models are missing here (see Table 1 of the same paper) - It will be good to distinguish Masked Language Modelling (MLM) and Autoregressive Token Generation (ATG).}

\subsection*{Chemical Language and Tokenization}
\label{sec:smiles_representation_tokenization}

The Simplified Molecular Input Line Entry System (SMILES) \cite{SMILES} provides a structured method for representing molecular structures as linear strings, making it well-suited for processing by large language models (LLMs). However, the effectiveness of these models depends heavily on how molecular structures are tokenized—a critical factor influencing both learning efficiency and predictive performance. Several tokenization strategies have been explored, each impacting model performance in distinct ways. In general, SMILES tokenization can be categorized into three main approaches. \textbf{Character-level tokenization} treats each character in a SMILES string as an independent token. While simple and easy to implement, it introduces challenges such as the incorrect splitting of multi-character atomic symbols—for instance, \textbf{Br (bromine)} being mistakenly split into \textbf{B} and \textbf{r}. \textbf{Atom-level tokenization} enhances interpretability and chemical consistency by ensuring that each token corresponds to a meaningful chemical entity, allowing models to better capture molecular substructures. This approach has been adopted by ChemBERTa \cite{Chithrananda2020ChemBERTaLS} and Chemformer \cite{Irwin_2022}, while ChemFM models \cite{cai2024foundation} employ a similar strategy but with a predetermined vocabulary of 266 tokens, compared to the 591-token vocabulary used in SmilesTokenizer.\footnote{\url{https://deepchem.readthedocs.io/en/2.4.0/api_reference/tokenizers.html}} \textbf{Motif-level tokenization} segments molecules into commonly occurring chemical substructures rather than individual atoms. This method can be guided by expert-defined rules or data-driven techniques such as byte pair encoding (BPE). Models like ChemBERTa-2 \cite{ahmad2022chemberta2chemicalfoundationmodels} integrate motif-level tokenization with molecular property prediction (MPP) tasks, improving predictive accuracy. Other models such as X-Mol \cite{XUE2022899} and BARTSmiles \cite{chilingaryan2022bartsmiles} leverage motif-based tokenization to create more structured molecular representations. Chemformer \cite{Irwin_2022}, which integrates both masked language modeling (MLM) and autoregressive token generation (ATG), benefits from this approach, achieving strong results in both molecular property prediction and molecule generation. The impact of tokenization strategies is evident in downstream tasks, where models that utilize more structured and chemically meaningful tokenization methods tend to exhibit better generalization and efficiency. For instance, Chemformer \cite{Irwin_2022} outperforms baseline models in molecular property prediction and generative tasks, while BARTSmiles \cite{chilingaryan2022bartsmiles} excels in reaction prediction and molecular design.

\subsection*{Target-aware Ligand Generation}

A major breakthrough in chemical generative models is their ability to design molecules tailored to specific protein targets. The development of new pharmaceuticals—from target identification to regulatory approval—is an extensive and costly process, typically spanning 12 years and costing approximately \$2.6 billion \cite{dhakal2022artificial}. To accelerate this pipeline, machine learning (ML) and deep learning (DL) have increasingly been employed to enhance computational drug discovery. While many computational methods generate bioactive molecules by leveraging known protein-ligand binding sites, their efficacy diminishes when binding sites are undefined or ambiguous. Binding site identification remains an open challenge, limiting the effectiveness of structure-informed approaches that rely solely on predefined binding pockets \cite{verkhivker2001binding}.
A recent advancement addressing this limitation is TargetVAE, a variational autoencoder introduced by Ngo and Hy \cite{ngo2024multimodal}, which employs multimodal protein representations—including amino acid sequences, three-dimensional structures, and residue-level graphs—to generate ligands without explicitly predefined binding sites. In contrast, our approach introduces a fully sequence-driven multimodal framework, integrating our protein Language Model (pLM) with our chemical Language Model (cLM). This enables the direct generation of ligands from protein sequences alone, eliminating the need for explicit structural or residue-level features. By leveraging protein sequence embeddings to guide chemical generation, our method streamlines the computational pipeline while maintaining strong target specificity. This approach significantly expands the scope of target-aware ligand generation, making it possible to design bioactive molecules even when binding sites are unknown or poorly defined.

\section{Methodology}
\label{sec:Methodology}

We introduce a novel approach leveraging our family of chemical Language Models (cLMs), specifically designed for drug-like and target-aware ligand generation. Unlike conventional molecular representations, our approach explicitly captures atomic-level interactions, providing a chemically meaningful and information-rich representation of molecules. By embedding molecules within a language that inherently models atomic interactions, we establish a precise and generative framework for designing compounds with desired pharmacological properties. Our contributions in this paper are threefold: (i) We develop a novel chemical language that encodes atomic interactions, enhancing both the representation and generation of drug-like molecules; (ii) We introduce a multi-omics modality integration framework by combining our chemical language model with a protein Language Model (Prot42 \cite{Prot42}), enabling target-aware ligand generation aligned with structural and functional properties of the target; and (iii) We conduct a comprehensive evaluation of our approach in terms of chemical validity, binding affinity prediction, and target specificity. By bridging chemical and protein Language Models, we advance toward a rational and data-driven drug design paradigm, unlocking new possibilities for highly specific therapeutic development. Our chemical language models (cLMs) were pre-trained on the extensive UniChem dataset \cite{unichem} using a decoder-only, autoregressive LLaMA-inspired architecture \cite{touvron2023llama}. To determine the optimal model size and hyperparameters, we employed maximal update parametrization ($\mu$P) [mup-mutransfer], leveraging a 38M-parameter proxy model. After fine-tuning hyperparameters through this proxy, we conducted full-scale pre-training, ensuring both scalability and efficiency. Evaluations demonstrated that a target tokens-per-parameter (TPP) ratio of 50 provided the optimal balance, achieving the lowest validation loss while maintaining computational efficiency. This approach optimizes performance while reducing computational overhead, setting a new benchmark for efficient and scalable chemical modeling.

\subsection{Pretraining Data and Tokenization}
\label{sec:Chem42Data}
The data preprocessing pipeline begins with the UniChem database, which contains small molecules as well as larger biomolecular entities, including nucleotides, carbohydrates, lipids, peptides, and chemically modified macromolecules. To standardize molecular representations, we extract molecular structures and convert them into their canonical SMILES format \cite{SMILES}. Following previous methodologies \cite{cai2024foundation}, SMILES enumeration is applied to enhance data diversity by generating multiple random variations of each molecule. This process yields four distinct dataset partitions, allowing us to train models of different sizes and evaluate their performance across various dataset configurations. For tokenization, we adopt an atom-level tokenization strategy, similar to that used in ChemFM \cite{cai2024foundation}, featuring a vocabulary of 268 tokens. This vocabulary includes periodic table elements, digits, SMILES-specific special characters, and start/end sequence markers. The dataset is tokenized with a context length of 512 tokens, ensuring effective representation of molecular sequences. The dataset statistics are presented in Table \ref{tab:datasets}.

\begin{table}[ht!]
\centering
\scriptsize
\begin{tabular}{l|c|c|c}
\hline

\textbf{Dataset Name} & \textbf{Val Size} & \textbf{Train Size} & \# \textbf{Nonpad Training Tokens}\\
\hline
% PubChem-Canonical & 2.6M & 115.6M & 5.8B\\
UniChem-Canonical & 7.65M & 188.4M & 9.5B\\
UniChem-Canonical+Random1 & 15.3M & 376.9M & 19.5B \\
UniChem-Canonical+Random3 & 15.3M & 565.4M & 29.5B \\
UniChem-Canonical+Random5 & 15.3M & 942.3M & 49.5B \\
\hline
\end{tabular}
\caption{Pretraining dataset specifications used to train Chem42 models.}
\label{tab:datasets}
\end{table}

\subsection{Model Architecture, Scaling-laws and Pretraining}
\label{sec:Architecture}

\begin{figure*}[ht!]
    \centering
    \includegraphics[width=.9\columnwidth]{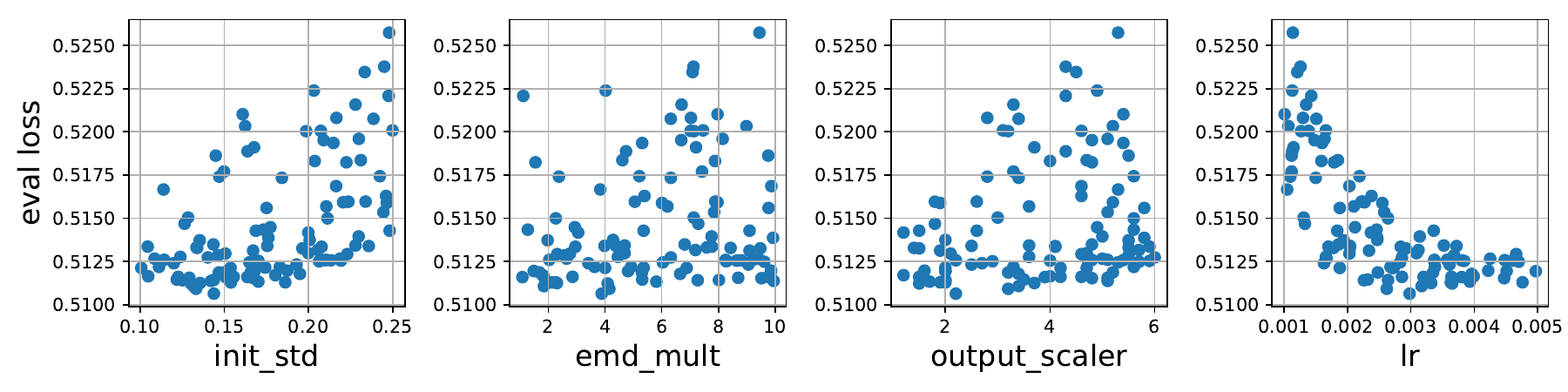}
    \caption{\(\mu\)P Hyperparameter Optimization. We perform a hyperparameter sweep consisting of 200 independent runs to optimize the standard deviation of parameter initializers, the embedding multiplier, the output layer logit multiplier, and the peak learning rate. The optimal hyperparameter values identified through this process are 0.1438, 3.88, 2.2, and $2.98 \times 10^{-3}$, respectively.}
    \label{fig:mup_sweep}
\end{figure*}

Chem42 is an autoregressive transformer decoder model based on the LLaMA architecture \cite{touvron2023llama}. To optimize hyperparameter selection for training, we employ maximal update parametrization (\(\mu\)P) \cite{mup-mutransfer}. Our hyperparameter tuning process is conducted using a smaller 38M-parameter proxy model. The results of our hyperparameter sweep are presented in Figure \ref{fig:mup_sweep}. After identifying the optimal hyperparameter configuration, we apply scaling laws to determine the optimal token-per-parameter (TPP) ratio for pre-training. We define five distinct compute budgets, ranging from \(5 \times 10^{18}\) to \(4.59 \times 10^{19}\), and systematically vary the model size and TPP for each budget. Subsequently, we pre-train these models and evaluate validation accuracy to identify the optimal TPP. For each training run, we utilize \(\mu\)Transfer to configure hyperparameters, employing a learning rate schedule that includes a linear warm-up followed by a 10× cosine decay. 

\begin{figure*}[ht!]
    \centering
    \includegraphics[width=.8\columnwidth]{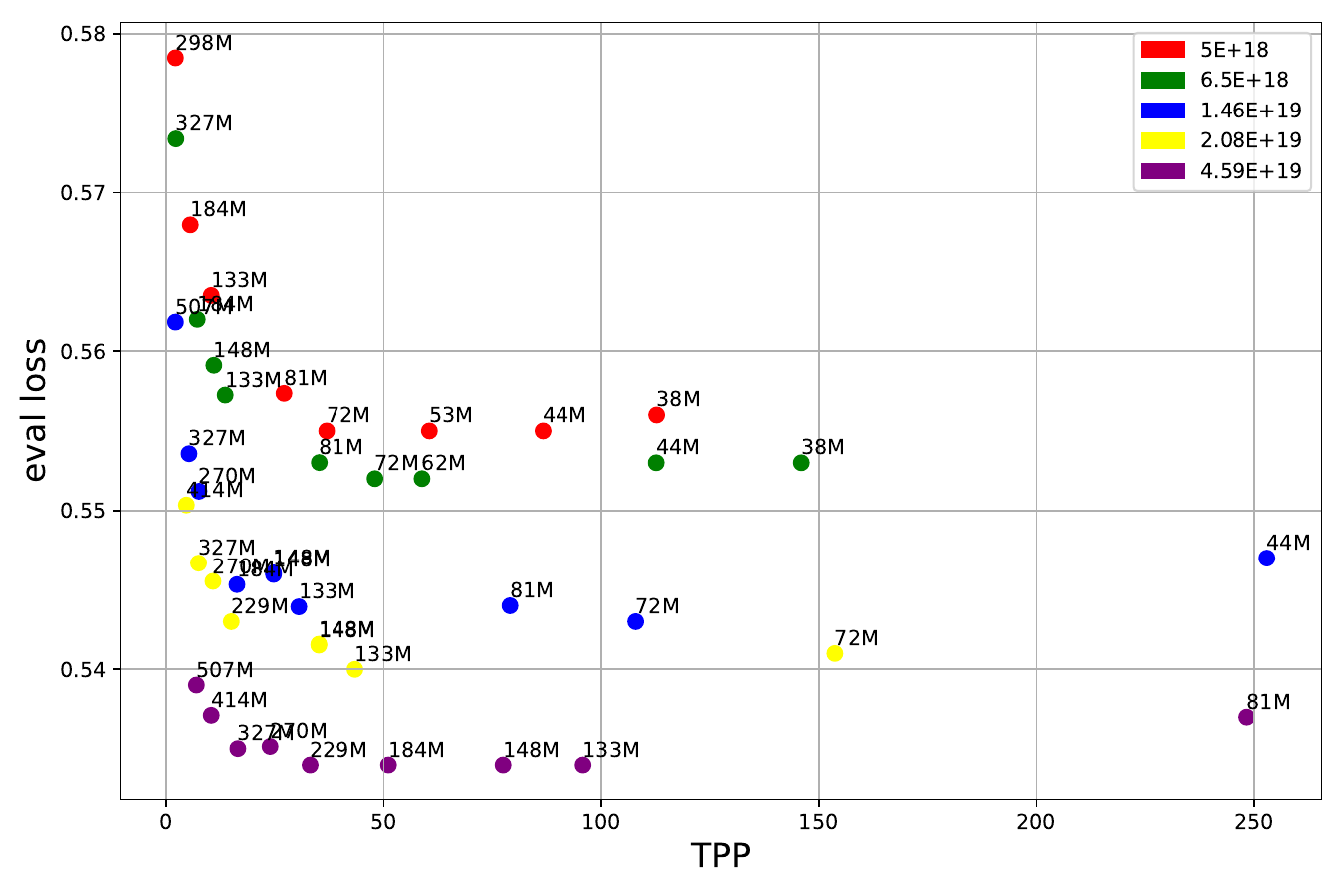}
    \caption{Scaling Law Experiments. We plot validation loss against token per parameter TPP. Colors indicate the compute budget measured in FLOPs, the needed floating-point operations.}
    \label{fig:scaling_law}
\end{figure*}

The results of our scaling law experiments are depicted in Figure \ref{fig:scaling_law}. The scaling laws suggest that a TPP of \textbf{50} is optimal for our pre-training process. We also present models and hyperparameters used in scaling law experiments in Table \ref{tab:scaling_law_models}.

We utilize Cerebras CS-2 for our training runs. The Cerebras CS-2 system is an AI accelerator that features 850,000 AI optimized compute cores, 40GB of on-chip SRAM, 20 PB/s memory bandwidth, and 220 PB/s interconnect \cite{cs2specs}. We present our production run model architecture and configuration in Table \ref{table:Chem4242-c-models-training-hyperparams}.

\begin{table}[ht!]
\centering
\scriptsize
\addtolength{\tabcolsep}{0pt}
\def\arraystretch{1.2}
    \begin{tabular}{l| c c c c}

            \textbf{Model} & \textbf{Chem42-190M} &\textbf{Chem42-387M} &  \textbf{Chem42-597M} &  \textbf{Chem42-1B} \\
            \shline
            Batch size & 512 & 512 & 512 & 512 \\
            Base Frequency & 10k & 10k & 10k & 10k \\
            Hidden size & 768 & 1088 & 1408 & 1792 \\
            \# of hidden layers & 24 & 24 & 24 & 24 \\
            \# of attention heads  & 16 & 16 & 16 & 32 \\
            \# KV heads & 4 & 4 & 4 & 8 \\
            Transformer FFN Dim. & 2816 & 4032 & 4480 & 6272 \\
        \hline
            Optimizer & AdamW & AdamW & AdamW & AdamW \\
            Betas  0.9, 0.95 & 0.9, 0.95 & 0.9, 0.95 & 0.9, 0.95\\
            Eps & 1e-8 & 1e-8 & 1e-8 & 1e-8\\
            Weight Decay & 0.1 & 0.1 & 0.1 & 0.1\\
            Max grad norm & 1 & 1 & 1 & 1\\
            Learning rate (Linear) & 0 to 5.9e-4 & 0 to 2.9e-4 & 0 to 1.9e-4 & 0 to 1.1e-4\\
            Iterations (Linear) & 0 to 1800 & 0 to 3700 & 0 to 5690 & 0 to 9650 \\
            Learning rate (Cosine) & 5.9e-4 to 5.9e-5 & 2.9e-4 to 2.9e-5 & 1.9e-4 to 1.9e-5 & 1.1e-4 to 1.1e-5 \\
            Iterations (Cosine) & 1800 to 36250 & 3700 to 74250 & 5690 to 113800 & 9650 to 193000 \\
        
    \end{tabular}
    \caption{Hyperparameters used for pretraining of the Chem42 models.}
    \label{table:Chem4242-c-models-training-hyperparams}
\end{table}

\subsection{Baseline Family of Models}
\label{sec:Basline}

To further compare Chem42 on downstream prediction tasks and generation tasks, we built a baseline family of models pretrained on the PubChem dataset (77M compounds) and using ChemBERTa-2 tokenization \cite{ahmad2022chemberta2chemicalfoundationmodels} which consists of a vocab size of \textbf{591 tokens} (larger compared to the vocab size used in Chem42 \cite{takase2024large} (\textbf{268 tokens})). Compared to the UniChem dataset described in Section.\ref{sec:Chem42Data}, PubChem consists of three primary databases: Substance, Compound, and BioAssay which makes it richer in biological activities of small molecules. We will report in Section.\ref{sec:Evaluations} and the Appendix quantitative results of Chem42 models and Baseline models.      

\begin{table}[ht!]
\centering
\scriptsize
\begin{tabular}{c|c|c|c|c|c|c|c}
%\hline

\textbf{Version} & \textbf{Model} & \textbf{Dataset} & \textbf{Smiles enumeration} & \textbf{Num sequences} & \textbf{Tokenizer} & \textbf{Vocabulary Size} & \textbf{TPP}\\
\hline

\multirow{3}{*}{Baseline} & 140M & Pubchem & Canonical & 115.6M & Motif Level & 591 & 50\\
& 507M & Pubchem & Canonical & 115.6M & Motif Level & 591 & 50\\
& 184M & Unichem & Canonical & 188.4M & Motif Level & 591 & 50\\
\hline

\multirow{4}{*}{Chem42} & 190M & Unichem & Canonical & 188.4M & Atom Level & 268 & 50\\
& 387M & Unichem & Canonical+1xRandom & 376.9M & Atom Level & 268 & 50\\
& 597M & Unichem & Canonical+2xRandom & 565.4M & Atom Level & 268 & 50\\
& 1002M & Unichem & Canonical+4xRandom & 942.3M & Atom Level & 268 & 50\\

\end{tabular}
\caption{Baseline vs Chem42 models.}
\label{tab:chem42_version_comparison}
\end{table}

\subsection{Protein-binding Ligand Generation}
\label{sec:Binders}

Given a dataset $\mathcal{D}$ of protein-ligand pairs, our objective is to generate ligand sequences $\hat{S}_l$ that exhibit high affinity toward a target protein. Let $(S_p, S_l) \in \mathcal{D}$ represent a protein-ligand pair, where $S_p = (s_1, s_2, ..., s_m)$ is a sequence of $m$ amino acid tokens, and $S_l = (l_1, l_2, ..., l_n)$ is a sequence of $n$ ligand tokens, each belonging to a predefined vocabulary $\mathcal{V}_l$. The protein model (Prot42 \cite{Prot42} $M_p$ encodes $S_p$ into a sequence of hidden representations $\mathbf{E}_p = (e_1, e_2, ..., e_m)$, where each $e_i \in \mathbb{R}^{2048}$ captures structural and functional properties of the protein. Similarly, the ligand model (Chem42) $M_l$ encodes $S_l$ into a sequence of latent embeddings $\mathbf{H}_l = (h_1, h_2, ..., h_n)$, where each $h_t \in \mathbb{R}^{1280}$ represents the hidden state of the chemical model. To integrate protein context into ligand generation, protein embeddings are projected into the hidden dimension of the ligand model using a learnable transformation $\mathbf{E}_p' = \theta_p \mathbf{E}_p$, where $\theta_p \in \mathbb{R}^{2048 \times 1280}$. These transformed protein embeddings are incorporated into ligand representations using a cross-attention mechanism. Attention scores are computed as 
$
A = \text{Softmax} \left( \frac{QK^\top}{\sqrt{1280}} \right)$,
where the query, key, and value projections are defined as 
$
Q = \theta_q \mathbf{H}_l, \quad K = \theta_k \mathbf{E}_p', \quad V = \theta_v \mathbf{E}_p',
$
with learnable weight matrices $\theta_q, \theta_k, \theta_v \in \mathbb{R}^{1280 \times 1280}$. The resulting protein-conditioned ligand representations are given by $\mathbf{C}_l = A V$, which are combined with the original ligand embeddings as $\mathbf{H}_l' = \mathbf{H}_l + \mathbf{C}_l$. 
These transformed representations $\mathbf{H}_l'$ are then passed through the CFM decoder layers. The generation of ligands follows an autoregressive process in which each token $\hat{l}_t$ is predicted based on previously generated tokens and protein-informed features. The probability distribution over the ligand vocabulary $\mathcal{V}_l$ is given by  $y_t = \text{Softmax}(\theta_h h_t' + \theta_c C_t),$
where $\theta_h, \theta_c$ are learnable parameters, and the next ligand token is selected as  
$\hat{l}_t = \arg\max_{j} (y_t^j).$
This process continues until a termination token is reached. 
 The generated ligand sequence $\hat{S}_l = (\hat{l}_1, \hat{l}_2, ..., \hat{l}_{\hat{n}})$ is then ranked using a scoring function $R(\hat{S}_l)$, which evaluates molecular validity, drug-likeness (QED) and synthetic accessibility (SA).
Thus, the entire protein-ligand generation process can be represented as:
$\hat{S}_l = M_l(\mathbf{E}_p', \mathbf{H}_l),$
where the parameters of this model are $\theta = \{\theta_p, \theta_q, \theta_k, \theta_v, \theta_h, \theta_c\}$. 

\begin{figure}[ht!]
    \centering
    \includegraphics[width=.8\columnwidth]{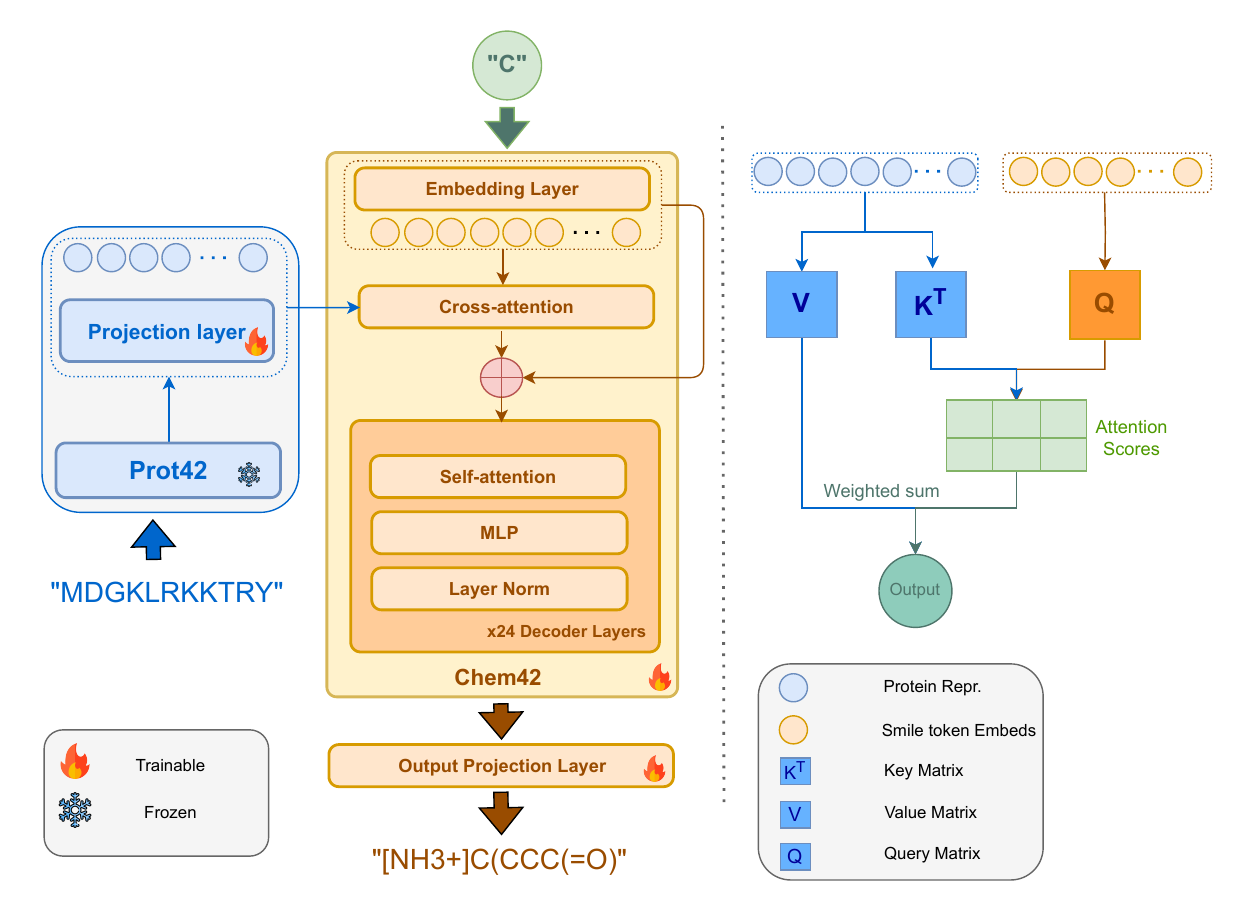}
    \caption{Overview of the our protein-ligand model architecture. 
       (Left) Our multi-omics pipeline integrates protein embeddings into the chemical model via cross-attention before passing through the decoder layers. The embeddings are first projected into the model’s hidden space.  
(Right) Cross-attention enables the model to compute interactions between chemical tokens and protein embeddings. After projection, protein embeddings serve as key-value pairs, while token embeddings act as queries. The attended representations are then combined with token inputs and processed by the decoder.}
    \label{fig:PLI}
\end{figure}

During inference, the protein embeddings \( \mathbf{E}_p \) are passed to $M_l$ , where the generation of the ligand sequence begins with a fixed token, ``C'' (see Figure \ref{fig:PLI}). This token serves as a controlled starting point for generating SMILES string. The ``C'' (Carbon) token represents a generic initialization that anchors the generation process, ensuring that the ligand sequence starts from a valid structure. By fixing the ``C'' token, the model can focus on generating the rest of the ligand sequence while keeping the protein context in mind. 

% To condition ligand generation on protein context, a cross-attention mechanism integrates \(\mathbf{E}_p\) with \(\mathbf{H}_l\), producing a contextualized ligand representation \(\mathbf{C}_l\).

\section{Experimental Results}
\label{sec:Evaluations}

\subsection{Molecular property prediction}

To assess the efficacy of our Chem42 models in downstream tasks, we evaluated them on two widely used benchmarks, Moleculenet\cite{wu2018moleculenet} and Admet\cite{huang2021therapeutics}. We adopted the same experimental settings and hyperparameters as specified in the ChemFM GitHub repository \footnote{\url{https://github.com/TheLuoFengLab/ChemFM}}, ensuring consistency across all tasks. This includes using the same random seed to maintain reproducibility. Tables \ref{tab:moleculenet_benchmark} and \ref{tab:admet_benchmark} present the results, comparing the performance of Chem42 against ChemFM. Our models outperform ChemFM on the majority of benchmarks while maintaining significantly fewer parameters, demonstrating the efficiency and effectiveness of our approach.

\paragraph{Evaluation on MoleculeNet Benchmark}

The Moleculenet\cite{wu2018moleculenet} benchmark comprises 12 different datasets with varying sample sizes across 5 different molecular properties and could be either a classification or regression prediction task. 
Table \ref{tab:moleculenet_benchmark} shows the comparison between ChemFM and our Chem42 models. Our Chem42 models outperform the ChemFM models on 10 of 12 tasks while having significantly fewer model parameters compared to ChemFM. On the BACE task, our best model has an ROC-AUC that is just 0.002 lower than ChemFM. 

\begin{table}[ht!]
    \centering
    \scriptsize
    \begin{tabular}{|c|c|c|c|c|c|c|c|c|c|c|c|c|}
        \hline
        \multirow{2}{*}{\textbf{Category}} & \multirow{2}{*}{\textbf{Dataset}} & \multirow{2}{*}{\textbf{Metric}} & \multicolumn{2}{c|}{\textbf{ChemFM} \cite{cai2024foundation}} & \multicolumn{4}{c|}{\textbf{Chem42 (Ours)}} \\
         
         &  &  & 1B & 3B & 190M & 387M & 597M & 1B \\
        
        \hline
        Pharmacokinetic & BBBP & ROC-AUC $\uparrow$ & 0.745 & 0.751 & 0.770 & \cb \textbf{0.775} & 0.763 & 0.762 \\
        
        \hline
        \multirow{4}{*}{Bioactivity} & BACE & ROC-AUC $\uparrow$ & 0.857 & \textbf{0.869} & 0.848 & 0.851 & 0.853 & \cb 0.867 \\
        & HIV & ROC-AUC $\uparrow$ & 0.785 & \textbf{0.807} & 0.774 & \cb 0.789 & 0.774 & 0.732 \\
        & MUV & ROC-AUC $\uparrow$ & 0.812 & - & 0.824 & 0.850 & \cb \textbf{0.853} & 0.712 \\
        & PCBA & PRC-AUC $\uparrow$ & 0.785 & 0.807 & \cb \textbf{0.892} & - & - & 0.886 \\
        
        \hline
        \multirow{3}{*}{Toxicity} & TOX21 & ROC-AUC $\uparrow$ & 0.863 & 0.869 & 0.836 & \cb \textbf{0.873} & 0.862 & 0.870 \\
        & SIDER & ROC-AUC $\uparrow$ & 0.702 & 0.709 & 0.693 & 0.698 & \cb \textbf{0.716} & 0.578 \\
        & CLINTOX & ROC-AUC $\uparrow$ & 0.899 & 0.918 & \cb \textbf{0.976} & 0.962 & 0.939 & 0.971 \\
        
        \hline
        \multirow{4}{*}{Physicochemical} & ESOL & RMSE $\downarrow$ & 0.529 & 0.516 & 0.480 & \cb \textbf{0.468} & 0.511 & 0.470 \\
        & FREESOLV & RMSE $\downarrow$ & 0.906 & 0.830 & 0.638 & \cb \textbf{0.582} & 0.673 & 1.023 \\
        & LIPO & RMSE $\downarrow$ & 0.547 & 0.545 & 0.568 & \cb \textbf{0.506} & 0.519 & 0.578\\
        
        \hline
        Molecular Binding & PDBBind Full & RMSE $\downarrow$ & 0.700 & 0.697 & 0.743 & 0.704 & 0.697 & \cb \textbf{0.674} \\
        \hline
        
    \end{tabular}
    \caption{Experimental Results on Moleculenet \cite{wu2018moleculenet} benchmark datasets for molecular property prediction following the experimental settings of \cite{cai2024foundation}. \textbf{Bold}: best performance; \colorbox{green!30}{Green} indicates best performance within the Chem42 models. }
    \label{tab:moleculenet_benchmark}
\end{table}

Table \ref{tab:extended_moleculenet_benchmark} in the Appendix compares Chem42 performances to the Baseline family of models described in Section\ref{sec:Basline} on MoleculeNet benchmark.      

\paragraph{Evaluation on ADMET Benchmark}

The ADMET\cite{huang2021therapeutics} benchmark comprises of 22 datasets of varying sizes across 5 molecular properties. Similar to Moleculenet, this benchmark has tasks split across classification and regression. Our Chem42 models outperform ChemFM on 18 of 22 tasks (table \ref{tab:admet_benchmark}) while having significantly fewer model parameters compared to ChemFM.

    \begin{table}[ht!]
    \centering
    \scriptsize
    \begin{tabular}{|c|c|c|c|c|c|c|c|c|c|c|c|c|}
        \hline
        \multirow{2}{*}{\textbf{Category}} & \multirow{2}{*}{\textbf{Dataset}} & \multirow{2}{*}{\textbf{Metric}} & \textbf{ChemFM} & \multicolumn{4}{c|}{\textbf{Chem42 (Ours)}} \\
         
         &  &  & 3B \cite{cai2024foundation} & 190M & 387M & 597M & 1B \\
        
        \hline
        \multirow{6}{*}{Absorption} & Caco2\_Wang & MAE $\downarrow$ & 0.322 & 0.296 & \cb\textbf{0.286} & 0.289 & \cb\textbf{0.286} \\
        & Bioavailability\_Ma & ROC-AUC $\uparrow$ & 0.715 & 0.658 & 0.725 & \cb\textbf{0.745} & 0.739 \\
        & Lipophilicity\_AstraZeneca & MAE $\downarrow$ & 0.460 & 0.493 & \cb\textbf{0.455} & 0.464 & 0.463 \\
        & Solubility\_AqSolDB & MAE $\downarrow$ & 0.725 & 0.717 & 0.712 & \cb\textbf{0.706} & 0.735 \\
        & HIA\_Hou & ROC-AUC $\uparrow$ & 0.984 & 0.993 & 0.991 & \cb\textbf{0.995} & \cb\textbf{0.995} \\
        & Pgp\_Broccatelli & ROC-AUC $\uparrow$ & \textbf{0.931} & 0.910 & 0.927 & \cb 0.930 & 0.855 \\
        
        \hline
        \multirow{3}{*}{Distribution} & BBB\_Martins & ROC-AUC $\uparrow$ & 0.908 & 0.919 & 0.932 & 0.932 & \cb\textbf{0.995} \\
        & PPBR\_AZ & MAE $\downarrow$ & 7.505 & 8.152 & \cb\textbf{7.479} & 7.677 & 7.662 \\
        & VDss\_Lombardo & Spearman $\uparrow$ & 0.662 & 0.702 & 0.711 & \cb\textbf{0.739} & 0.694 \\
        
        \hline
        \multirow{6}{*}{Metabolism} & CYP2C9\_Veith & PRC-AUC $\uparrow$ & \textbf{0.788} & 0.744 & \cb 0.768 & 0.767 & 0.765 \\
        & CYP2D6\_Veith & PRC-AUC $\uparrow$ & \textbf{0.704} & 0.667 & \cb 0.684 & 0.678 & 0.680 \\
        & CYP3A4\_Veith & PRC-AUC $\uparrow$ & 0.878 & \cb \textbf{0.881} & 0.876 & \cb \textbf{0.881} & 0.875 \\
        & CYP2C9\_Substrate\_CarbonMangels & PRC-AUC $\uparrow$ & 0.414 & 0.514 & 0.501 & 0.509 & \cb \textbf{0.548} \\
        & CYP2D6\_Substrate\_CarbonMangels & PRC-AUC $\uparrow$ & 0.739 & 0.718 & 0.702 & \cb\textbf{0.740} & 0.522 \\
        & CYP3A4\_Substrate\_CarbonMangels & PRC-AUC $\uparrow$ & 0.654 & 0.620 & \cb\textbf{0.656} & 0.637 & 0.595 \\
        
        \hline
        \multirow{3}{*}{Excretion} & Half\_Life\_Obach & Spearman $\uparrow$ & 0.551 & 0.596 & 0.613 & 0.595 & \cb\textbf{0.615} \\
        & Clearance\_Hepatocyte\_AZ & Spearman $\uparrow$ & 0.495 & 0.424 & 0.459 & 0.490 & \cb\textbf{0.499} \\
        & Clearance\_Microsome\_AZ & Spearman $\uparrow$ & 0.611 & 0.620 & 0.614 & 0.622 & \cb\textbf{0.633} \\

        \hline
        \multirow{4}{*}{Toxicity} & LD50\_Zhu & MAE $\downarrow$ & 0.541 & 0.530 & \cb\textbf{0.520} & 0.522 & 0.524 \\
        & hERG & ROC-AUC $\uparrow$ & 0.848 & \cb\textbf{0.859} & 0.847 & 0.815 & 0.830 \\
        & AMES & ROC-AUC $\uparrow$ & \textbf{0.854} & 0.833 & \cb 0.850 & \cb 0.850 & 0.846 \\
        & DILI & ROC-AUC $\uparrow$ & 0.920 & \cb\textbf{0.929} & 0.887 & 0.896 & 0.923 \\
        
        \hline
    \end{tabular}
    \caption{Experimental Results on ADMET \cite{huang2021therapeutics} benchmark for property prediction following the experimental setting of \cite{cai2024foundation}. \textbf{Bold}: best performance; \colorbox{green!30}{Green} indicates best performance within the Chem42 models. }
    \label{tab:admet_benchmark}
\end{table}

Table \ref{tab:extended_admet_benchmark} in the Appendix compares Chem42 performances to the Baseline family of models described in Section\ref{sec:Basline} on ADMET benchmark. 

\subsection{Unconditional molecular generation}
\label{sec:unconditionl_gen}

We model unconditional generation as a sequence completion problem. Provided an input sequence $S_{in}$, we ask the model to complete the sequence by predicting the next token until the end of sequence token $[EOS]$ is encountered. We generate multiple such smiles strings and evaluate the generated strings using 2 metrics: Validity and Uniqueness. We define Validity score as follows, $Validity = |S_{valid}| / |S_{gen}|$, where $S_{valid}$ is the set of all valid smiles strings. Validity of a smiles string is found via the RDKit library \cite{bento2020open}. $S_{gen}$ is the complete set of all generated smiles strings. We define Uniqueness score as follows, $Uniqueness = |S_{unique}| / |S_{gen}|$, where $S_{unique}$ is the set of unique smiles strings. We conducted two experiments for unconditional prediction: (1) predicting the second half of the Ozempic molecule and (2) generating a new SMILES string starting from a single carbon atom. These tasks evaluate how the model's performance varies under different generation constraints. In the second task, where generation starts with a single carbon atom (\textbf{C}) and continues until an end-of-sequence ($[EOS]$) token is reached, the model operates without restrictions, increasing the likelihood of generating valid SMILES string. Conversely, in the first task, predicting the second half of Ozempic imposes a length constraint, making it more challenging to generate valid SMILES. Using a temperature of 1, we generate 1,000 SMILES strings for each task and evaluate their validity and uniqueness. In the first task, the initial half of the Ozempic SMILES string (comprising 326 tokens out of 653) is provided as input, and the model predicts the remaining tokens until an $[EOS]$ token appears. In the second task, we begin with \textbf{C} as input and generate tokens up to a maximum sequence length of 512. In both cases, generation stops if the $[EOS]$ token is predicted or the model reaches its maximum sequence length.

\begin{table}[ht!]
\centering
\scriptsize
\begin{tabular}{|c|c|c|l|c|c|c|}
\hline

\textbf{Input Molecule} & \textbf{Tokens to generate} & \textbf{Generated mol.} & \textbf{Model} & \textbf{Uniqueness} $\uparrow$ & \textbf{Validity} $\uparrow$ \\
\hline

\multirow{5}{*}{Ozempic} & 326 & 1000 & ChemFM 3B & 98.60 & 79.30 \\ \cline{2-6}
& 326 & 1000 & Baseline 507M & 91.80 & 75.70 \\ 
& 326 & 1000 & Chem42 387M & 97.30 & 89.30 \\ 
& 326 & 1000 & Chem42 597M & \cb\textbf{98.90} & 66.10 \\ 
& 326 & 1000 & Chem42 1B & 94.40 & \cb\textbf{91.60} \\ 
\hline 

\multirow{6}{*}{Single Carbon Atom 'C'} & $<=512$ & 1000 & ChemFM 3B & \textbf{99.80} & 100 \\ \cline{2-6}
& $<=512$ & 1000 & Baseline 507M & \cb 99.70 & 100 \\ 
& $<=512$ & 1000 & Chem42 387M & 99.60 & 100 \\ 
& $<=512$ & 1000 & Chem42 597M & \cb 99.70 & 100 \\ 
& $<=512$ & 1000 & Chem42 1B & 99.60 & 100 \\

\hline
\end{tabular}
\caption{Unconditional generation results. All numbers presented are percentage values. \textbf{Bold}: best performance; \colorbox{green!30}{Green} indicates best performance within the Chem42 models.}
\label{tab:unconditional_gen}
\end{table}

Table \ref{tab:unconditional_gen} compares the unconditional generation capabilities of our Chem42 models with ChemFM. Starting prediction from a single carbon atom, all models have a comparable performance thereby not providing us with any valuable insights into the generation capabilities of the models. Predicting the 2nd half of Ozempic is a more challenging task, and our models achieve a higher uniqueness score and comparable validity score while being significantly smaller than ChemFM models. 

\subsection{Evaluation on conditional molecule generation}
Conditional molecule generation plays a crucial role in designing molecules that satisfy specific property criteria or incorporate predefined scaffold structures. To address this, we fine-tuned one of our best performing Chem42 model - 597M on the GuacaMol dataset for property-based generation. The model was trained to optimize four continuous molecular properties: octanol-water partition coefficient (logP), synthetic accessibility score (SAS), topological polar surface area (TPSA), and quantitative estimate of drug-likeness (QED). For property-based generation, we evaluated the model trained on the GuacaMol dataset by generating 10,000 molecules per property combination and assessing validity, uniqueness, novelty, and mean absolute deviation (MAD) between the conditioned and computed properties. Our model demonstrated superior performance across all properties, achieving on average improvements of 0.017 in validity compared to MolGPT and 0.009 compared to ChemFM (Table \ref{tab:guacamol_benchmark}). Our model had a performance comparable or slightly worse than ChemFM on the other performance metrics while having a model size approximately 6 times smaller than ChemFM. Notably, we achieved the highest validity scores in all conditions, ranging from 0.981 to 0.996, surpassing both ChemFM-3B and MolGPT. For single-property generation, our Chem42 model outperformed its counterparts in logP (0.989), TPSA (0.989), SAS (0.996), and QED (0.993) in terms of validity, while also achieving the lowest mean absolute deviation (MAD) in SAS (0.107) and QED (0.043) (Table \ref{tab:guacamol_benchmark}).
In multi-property settings, our Chem42 maintained competitive performance, achieving the highest validity in all combinations, with values ranging from 0.981 to 0.988. We also recorded the lowest Mean Average Deviation (MAD) in SAS for all multi-property conditions, with a minimum of 0.111 (Table \ref{tab:guacamol_benchmark}).

\begin{table}[ht!]
    \centering
    \scriptsize
    \begin{tabular}{|c|l|c|c|c|c|}
        \hline
        \textbf{Property} & \textbf{Model} & \textbf{Validity $\uparrow$} & \textbf{Uniqueness $\uparrow$} & \textbf{Novelty $\uparrow$} & \textbf{MAD $\downarrow$} \\
        
        \hline 
        \multirow{3}{*}{logP} & MolGPT & 0.971 & 0.998 & 0.977 & 0.230 \\
        & ChemFM - 3B & 0.981 & \textbf{1.000} & \textbf{0.985} & \textbf{0.182} \\
        & Chem42 - 597M &\textbf{0.989} & 0.998 & 0.924 & 0.202 \\

        \hline 
        \multirow{3}{*}{TPSA} & MolGPT & 0.971 & 0.997 & 0.975 & 3.562 \\
        & ChemFM - 3B & 0.979 & \textbf{0.999} & \textbf{0.984} & \textbf{2.466} \\
        & Chem42 - 597M & \textbf{0.989} & 0.997 & 0.918 & 3.409 \\

        \hline 
        \multirow{3}{*}{SAS} & MolGPT & 0.978 & 0.996 & 0.966 & 0.133 \\
        & ChemFM - 3B & 0.986 & \textbf{0.999} & \textbf{0.971} & 0.126 \\
        & Chem42 - 597M & \textbf{0.996} & 0.995 & 0.897 & \textbf{0.107} \\

        \hline 
        \multirow{3}{*}{QED} & MolGPT & 0.974 & 0.997 & 0.968 & 0.056 \\
        & ChemFM - 3B & 0.982 & \textbf{1.000} & \textbf{0.980} & 0.045 \\
        & Chem42 - 597M & \textbf{0.993} & 0.999 & 0.899 & \textbf{0.043} \\

        \hline 
        \multirow{3}{*}{SAS + logP} & MolGPT & 0.972 & 0.991 & 0.983 & 0.147/ 0.253 \\
        & ChemFM - 3B & 0.980 & \textbf{0.995} & \textbf{0.985} & 0.137/ \textbf{0.195} \\
        & Chem42 - 597M & \textbf{0.987} & 0.983 & 0.934 & \textbf{0.111} / 0.228 \\

        \hline 
        \multirow{3}{*}{SAS + TPSA} & MolGPT & 0.971 & 0.988 & 0.984 & 0.155/ 3.785 \\
        & ChemFM - 3B & 0.980 & \textbf{0.991} & \textbf{0.985} & 0.138/ \textbf{2.659} \\
        & Chem42 - 597M & \textbf{0.988} & 0.975 & 0.936 & \textbf{0.111} / 3.597 \\

        \hline 
        \multirow{3}{*}{TPSA + logP} & MolGPT & 0.964 & 0.994 & 0.989 & 3.715/ 0.243 \\
        & ChemFM - 3B & 0.973 & \textbf{0.997} & \textbf{0.992} & \textbf{2.415}/ \textbf{0.184} \\
        & Chem42 - 597M & \textbf{0.984} & 0.990 & 0.957 & 3.292 / 0.211 \\

        \hline 
        \multirow{3}{*}{TPSA + logP + SAS} & MolGPT & 0.972 & 0.969 & 0.988 & 3.797/ 0.268/ 0.180 \\
        & ChemFM - 3B & 0.975 & \textbf{0.971} & \textbf{0.989} & \textbf{2.289}/ \textbf{0.191}/ 0.166 \\
        & Chem42 - 597M & \textbf{0.981} & 0.916 & 0.965 & 3.015/ 0.212/ \textbf{0.131} \\
        
        \hline
        
    \end{tabular}
    \caption{Experimental Results on GuacaMol dataset \cite{brown2019guacamol} for conditional generation following the experimental settings of \cite{cai2024foundation}. \textbf{Bold}: best performance}
    \label{tab:guacamol_benchmark}
\end{table}

% \subsection{Evaluation on Reaction Prediction}
% We fine-tuned our models for both reaction synthesis and retro-synthesis using the USPTO-FULL \cite{uspto_full}, USPTO-MIT \cite{uspto_mit}, and USPTO-50K \cite{uspto_50k} datasets. These datasets comprise organic chemical reactions extracted through text-mining from United States patents. To enhance the robustness and generalization capability of our models, we employed SMILES enumeration as a data augmentation strategy. Finally, we benchmarked our models against existing approaches and report the comparative results in Table \ref{tab:reaction_prediction}. 
\subsection{Evaluation on Reaction Prediction}
We fine-tuned our models for both reaction synthesis and retro-synthesis using 
the USPTO-FULL~\cite{uspto_full}, USPTO-MIT~\cite{uspto_mit}, and USPTO-50K~\cite{uspto_50k} datasets. These datasets consist of organic chemical reactions extracted through text mining from United States patents. To improve model robustness and generalization, we employed SMILES enumeration as a data augmentation strategy. Finally, we benchmarked our models against existing approaches and present the comparative results in Table~\ref{tab:reaction_prediction}.

\begin{table}[ht!]
    \centering
    \scriptsize
    \begin{tabular}{|c|c|c|c|c|l|c|c|c|}
        \hline
        \textbf{Task} & \textbf{Dataset} & \multicolumn{3}{c|}{\textbf{\# Samples}(Test/ Val/ Train)} & \textbf{Model} & \textbf{Top-1 $\uparrow$} & \textbf{Top-3 $\uparrow$} & \textbf{Top-5 $\uparrow$} \\
        
        \hline 
        \multirow{4}{*}{Synthesis} & \multirow{4}{*}{USPTO-MIT} & \multirow{4}{*}{40,265} & \multirow{4}{*}{30,182} & \multirow{4}{*}{411,685} & ChemFM - 3B & \textbf{90.5} & \textbf{95.7} & \textbf{96.6}\\
        & &  & & & Chem42 - 597M & 90.3 & \textbf{95.7} & \textbf{96.6}\\
        & &  & & & AT & \underline{90.4} & - & 96.5\\
        & &  & & & R-SMILES & 90.0 & 95.6 & 96.4\\

        \hline
        \multirow{15}{*}{Retro-synthesis} & \multirow{6}{*}{USPTO-50K} & \multirow{6}{*}{5,007} & \multirow{6}{*}{5,001} & \multirow{6}{*}{40,003} & ChemFM-3B & \underline{58.0} & \textbf{80.0} & \textbf{86.3} \\
        & & & & & Chem42 - 597M & \textbf{58.7} & \underline{79.5} & 83.6\\
        & & & & & AT & 53.5 & - & 81.0\\
        & &  & & & R-SMILES & 56.0 & 79.0 & \underline{86.1}\\
        & &  & & & RetroXpert & 50.4 & 61.1 & 62.3\\
        & &  & & & G$^{2}$Retro & 54.1 & 74.1 & 81.2\\
        \cline{2-9}
        
        & \multirow{4}{*}{USPTO-MIT} & \multirow{4}{*}{40,265} & \multirow{4}{*}{30,182} & \multirow{4}{*}{411,685} & ChemFM-3B & \underline{61.6} & \underline{78.7} & \textbf{83.0} \\
        & & & & & Chem42-597M & \textbf{62.4} & \textbf{78.8} & 82.6 \\
        & &  & & & R-SMILES & 60.3 & 77.9 & \underline{82.8}\\
        & &  & & & RetroTRAE & 58.3 & - & -\\

        \cline{2-9}
        & \multirow{5}{*}{USPTO-Full} & \multirow{5}{*}{96,023} & \multirow{5}{*}{96,071} & \multirow{5}{*}{768,630} & ChemFM-3B & \textbf{51.7} & 68.0 & \textbf{72.5} \\
        & & & & & Chem42-597M & \textbf{51.7} & \textbf{68.1} & \underline{72.2}\\
        & & & & & AT & 46.2 & - & 73.3\\
        & &  & & & R-SMILES & 48.9 & 66.6 & 72.0\\
        & &  & & & RetroXpert & 49.4 & 63.6 & 67.6\\

        \hline

    \end{tabular}
    \caption{Experimental results on the USPTO benchmarks. \textbf{Bold}: best performance and underlined values represent the second best performance.}
    \label{tab:reaction_prediction}
\end{table}

Empirical evaluation demonstrates that our Chem42 models achieve performance comparable to the state-of-the-art ChemFM 3B model, despite having six times fewer parameters. Consequently reduced model size enhances deployment efficiency for downstream applications. Furthermore, our findings validates our scaling law experiments, indicating that a token-to-parameter ratio of 50 TPP represents an optimal balance between model capacity and efficiency.

\subsection{Target-aware Ligand Generation}

This task focuses on generating drug-like molecules that can bind to target proteins with unknown binding sites. We used the PDBBind 2020 training set\footnote{\url{https://huggingface.co/datasets/jglaser/binding_affinity}}, which contains more than 16,000 unique protein-ligand pairs. For docking simulations and computing the binding affinity, we used AutoDock-Vina \cite{trott2010autodock}. Our approach is evaluated on eight target proteins; 

\begin{itemize}
    \item 1IEP\footnote{\url{https://www.rcsb.org/structure/1IEP}} (the kinase domain of \textit{c-Abl}) : The kinase domain of \textit{c-Abl} is a catalytic domain responsible for its tyrosine kinase activity. It consists of an N-terminal lobe primarily composed of $\beta$-sheets and a C-terminal lobe dominated by $\alpha$-helices, connected by a flexible hinge region. The ATP-binding site is located in the cleft between these lobes, facilitating phosphate transfer to substrate proteins. Regulation of \textit{c-Abl} activity involves autoinhibitory interactions and phosphorylation events that modulate its conformational state, influencing its role in signal transduction, cell proliferation, and cytoskeletal dynamics.

    \item 2RGP\footnote{\url{https://www.rcsb.org/structure/2rgp}} (EGFR (Epidermal Growth Factor Receptor)): The kinase domain of ErbB-2 (HER2) and EGFR is a tyrosine kinase domain that plays a central role in receptor dimerization and signal transduction. It consists of an N-lobe and C-lobe, with an ATP-binding cleft and an activation loop that regulates catalytic activity. In EGFR (ErbB-1), ligand binding induces dimerization, leading to autophosphorylation and activation of downstream signaling pathways like MAPK and PI3K-Akt, which drive cell proliferation and survival. HER2 (ErbB-2), though ligand-independent, is a potent signaling receptor due to its ability to form heterodimers with other ErbB family members. Mutations or overexpression of these kinases are associated with various cancers, making them \textbf{key targets for cancer therapies} like tyrosine kinase inhibitors (TKIs) and monoclonal antibodies.
    
    \item 3EML\footnote{\url{https://www.rcsb.org/structure/3EML}} (Human A2A Adenosine Receptor) : The human A$_{2A}$ adenosine receptor (A$_{2A}$AR) is a G protein-coupled receptor (GPCR) that belongs to the adenosine receptor family. It is primarily expressed in the central nervous system, cardiovascular system, and immune cells. A$_{2A}$AR plays a key role in modulating neurotransmission, vasodilation, and immune response by binding adenosine and activating intracellular signaling pathways via G$_s$ proteins, leading to increased cyclic AMP (cAMP) production. Due to its involvement in neurological disorders, inflammation, and cardiovascular regulation, A$_{2A}$AR is a significant therapeutic target in conditions such as Parkinson’s disease and ischemia.
    
    \item 3NY8\footnote{\url{https://www.rcsb.org/structure/3ny8}} (the human $\beta_2$-adrenergic receptor): The \textbf{human $\beta_2$-adrenergic receptor} ($\beta_2$AR) is a \textbf{G protein-coupled receptor (GPCR)} that plays a key role in mediating the effects of catecholamines, particularly \textit{epinephrine} and \textit{norepinephrine}. Unlike tyrosine kinases, $\beta_2$AR lacks a kinase domain and instead contains a \textbf{seven-transmembrane domain} characteristic of GPCRs. Upon ligand binding, $\beta_2$AR primarily couples to \textbf{$G_s$ proteins}, leading to the activation of \textit{adenylyl cyclase}, increased intracellular \textbf{cyclic AMP (cAMP)} levels, and subsequent activation of \textbf{protein kinase A (PKA)}. This signaling pathway regulates smooth muscle relaxation, bronchodilation, and cardiac function. $\beta_2$AR is a major pharmacological target for \textit{bronchodilators} in asthma and chronic obstructive pulmonary disease (COPD), as well as for cardiovascular therapeutics.
 
    \item 4RLU \footnote{\url{https://www.rcsb.org/structure/4rlu}} ((3R)-hydroxyacyl-ACP dehydratase HadAB hetero-dimer from Mycobacterium tuberculosis): The (3R)-hydroxyacyl-ACP dehydratase \textit{HadAB} is a heterodimeric enzyme involved in the fatty acid elongation cycle in \textit{Mycobacterium tuberculosis}. It consists of the HadA and HadB subunits, which function cooperatively to catalyze the dehydration of (3R)-hydroxyacyl-acyl carrier protein (ACP) intermediates during the biosynthesis of mycolic acids, essential components of the mycobacterial cell wall. Structural and functional analyses suggest that HadA provides catalytic activity, while HadB stabilizes the enzyme-substrate complex, ensuring efficient mycolate biosynthesis and bacterial survival.
 
    \item 4UNN\footnote{\url{https://www.rcsb.org/structure/4UNN}} (M. tuberculosis thymidylate kinase (Mtb TMK)): \textit{Mycobacterium tuberculosis} thymidylate kinase (Mtb TMK) is an essential enzyme in the nucleotide biosynthesis pathway, catalyzing the phosphorylation of dTMP to dTDP using ATP as a phosphate donor. It plays a crucial role in DNA replication and repair, making it a potential target for anti-tubercular drug development. Structurally, Mtb TMK consists of a conserved P-loop, which is critical for nucleotide binding, and a lid domain that undergoes conformational changes during catalysis. Its specificity and catalytic efficiency are regulated by interactions with substrates and cofactors.

    \item 5MO4\footnote{\url{https://www.rcsb.org/structure/5MO4}} (ABL1 kinase (T334I D382N)): The \textit{ABL1} kinase is a non-receptor tyrosine kinase involved in cell signaling, proliferation, and cytoskeletal dynamics. The T334I\_D382N double mutation affects the kinase domain, potentially altering its structural conformation and enzymatic activity. Threonine 334 to isoleucine (T334I) substitution may influence ATP binding or substrate interactions, while aspartate 382 to asparagine (D382N) could impact the regulatory mechanisms of the kinase. These mutations may play a role in resistance to tyrosine kinase inhibitors (TKIs) and altered oncogenic potential in diseases such as chronic myeloid leukemia (CML) and acute lymphoblastic leukemia (ALL).

    \item 7L11 \footnote{\url{https://www.rcsb.org/structure/7L11}} (Main Protease of SARS-CoV-2): The Main Protease (Mpro) of SARS-CoV-2, also known as 3C-like protease (3CLpro), is a key enzyme in the viral replication cycle. It plays a crucial role in processing the polyproteins translated from the viral RNA by cleaving them at specific sites to generate functional proteins essential for virus maturation. Due to its indispensable role in viral replication and its absence in humans, Mpro is a prime target for antiviral drug development against COVID-19.

\end{itemize}
\begin{figure}[ht!]
    \centering
    \includegraphics[width=.9\linewidth]{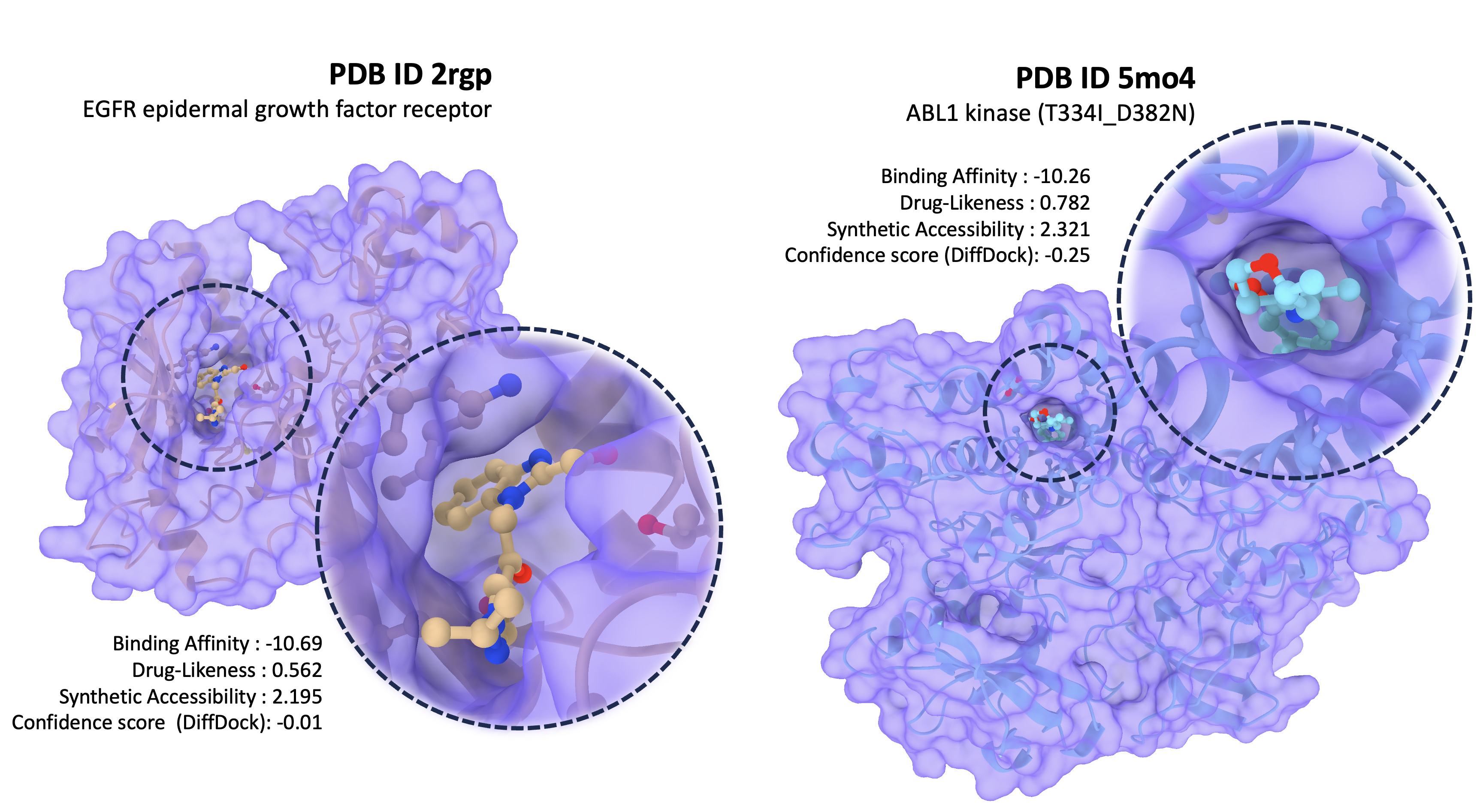}
    \includegraphics[width=.9\linewidth]{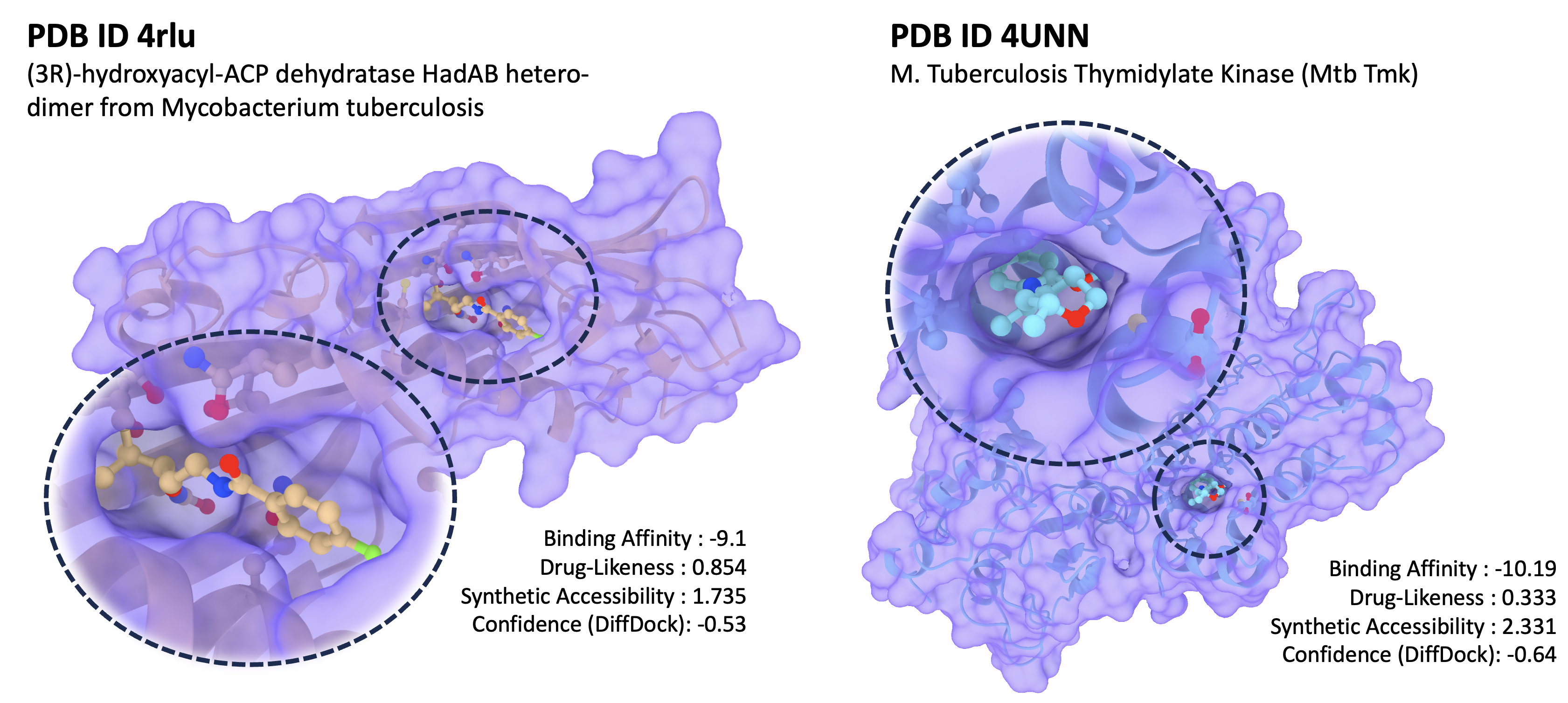}
    \caption{Examples of generated Ligands for protein targets 2rgp, 5mo4 4rlu and 4unn. Docking results with metrics (QED, SA, BA and Confidence score on pose estimation output of DiffDock-L \cite{corso2024deep}).}
    \label{fig:targets}
\end{figure}

These targets were not part of the model's training data. For each target protein, we generate 500 SMILES sequences, evaluating their drug-likeness (QED), synthetic accessibility (SA), both computed using RDKit \cite{bento2020open}. From these, we select the best molecule based on the properties mentioned, perform docking simulations with AutoDock Vina \cite{trott2010autodock}, and finally get the binding affinities. Table \ref{tab:PLI_metrics_comparison}  presents a comparative analysis of our multi-omics modality protein-ligand generation model against PMN \cite{ngo2024multimodal} across multiple target proteins. While binding affinities are generally comparable between the two models—with each showing strengths on different targets—our models consistently achieve higher QED and lower SA scores, demonstrating improved molecular quality and ease of synthesis-key factors for successful downstream drug development.
Figure \ref{fig:targets} visualizes some of our generated ligands.

% Our models consistently outperform PMN in QED and SA, demonstrating improved molecular quality and ease of synthesis. 
% Note that we compare absolute binding affinity values, as their magnitude directly reflects interaction strength, ensuring a consistent assessment of binding stability across different targets. 

\begin{table}[ht]
\centering
\caption{Performance comparison between PMN, \textbf{Chem42 (507M)}, and \textbf{Chem42 (597M)} on eight protein targets. Higher QED and more negative BA are desirable; lower SA indicates easier synthesis. \textbf{Bold} indicates the best performance per metric per target.}
\begin{adjustbox}{max width=\textwidth}
\begin{tabular}{@{}lccccccccc@{}}
\toprule
\multirow{2}{*}{\textbf{Target}} &
\multicolumn{3}{c}{\textbf{QED} ($\uparrow$)} &
\multicolumn{3}{c}{\textbf{SA} ($\downarrow$)} &
\multicolumn{3}{c}{\textbf{BA} ($\downarrow$) (kcal/mol)} \\
\cmidrule(lr){2-4} \cmidrule(lr){5-7} \cmidrule(lr){8-10}
 & PMN & Chem42 (507M) & Chem42 (597M) & PMN & Chem42 (507M) & Chem42 (597M) & PMN & Chem42 (507M) & Chem42 (597M) \\ \midrule
1iep  & 0.322 & \textbf{0.613} & 0.360 & 7.609 & 4.125 & \textbf{3.152} & -9.946 & \textbf{-10.860} & -10.35 \\
2rgp  & 0.428 & 0.526 & \textbf{0.562} & 3.391 & \textbf{2.195} & 2.334 & \textbf{-11.936} & -10.690 & -10.10 \\
3eml  & 0.584 & \textbf{0.624} & 0.534 & 7.268 & \textbf{1.895} & 3.271 & \textbf{-25.939} & -10.590 & -10.40 \\
3ny8  & \textbf{0.807} & 0.383 & 0.397 & 5.990 & \textbf{2.583} & 3.250 & -11.257 & \textbf{-12.070} & -10.70 \\
4rlu  & 0.479 & \textbf{0.854} & 0.803 & 2.979 & \textbf{1.735} & 1.765 & \textbf{-11.250} & -9.100 & -9.80 \\
4unn  & 0.161 & 0.333 & \textbf{0.463} & \textbf{2.323} & 2.331 & 3.628 & \textbf{-10.752} & -10.190 & -10.40 \\
5mo4  & 0.432 & \textbf{0.782} & 0.415 & 6.330 & \textbf{2.321} & 2.753 & \textbf{-11.812} & -10.260 & -10.23 \\
7l11  & 0.136 & 0.513 & \textbf{0.645} & 7.912 & 2.978 & \textbf{2.540} & \textbf{-11.220} & -7.580 & -8.43 \\
\bottomrule
\end{tabular}
\end{adjustbox}
\label{tab:PLI_metrics_comparison}
\end{table}

\section{Conclusion}
\label{sec:Conclusion}

In this work, we introduced Chem42, a new family of chemical Language Models (CLMs) designed for target-aware ligand generation by integrating atomic interaction modeling with protein Language Models (Prot42). Our approach addresses a critical limitation of existing molecular generative models by incorporating target-specific information through a multimodal framework, enabling the generation of ligands that are both chemically valid and biologically relevant. Through extensive evaluation, we demonstrated that Chem42 outperforms existing CLMs in terms of molecular validity, target specificity, and is comparable in predicting binding affinity. The ability of our model to capture atomic-level interactions enhances the precision of ligand design, while integration with protein embeddings ensures context-aware molecular generation, even in cases where binding sites are not well defined. Our experimental results validate the effectiveness of this approach, showing improved drug-likeness (QED), synthetic accessibility (SA), and on par with binding affinity (BA) in comparison to baseline methods.

The broader impact of this work extends beyond ligand generation. By bridging chemical and protein representations, we have paved the way for more accurate and efficient AI-driven drug discovery pipelines. Future directions include further scaling of the model, optimizing protein-ligand docking accuracy, and extending multimodal learning to incorporate structural and biochemical constraints. By integrating chemistry-aware and biology-aware representations, our work advances toward a data-driven, rational approach to drug design, ultimately accelerating the discovery of novel therapeutics with high specificity and efficacy.

\section*{Acknowledgment}
The authors are thankful to the AutoDock Vina team \cite{trott2010autodock}, a state-of-the-art model for molecular docking. All ligand-protein docking results reported in this paper are obtained using this tool. All protein visualizations reported in this paper were obtained using the UCSF ChimeraX software.   

\bibliographystyle{alpha}
\bibliography{main}

\newpage
\section*{Appendix}

\begin{table}[ht!]
    \centering
    \scriptsize
    \begin{tabular}{|c|c|c|c|c|c|c|c|c|c|c|c|c|c|}
        \hline
        \multirow{2}{*}{\textbf{Category}} & \multirow{2}{*}{\textbf{Dataset}} & \multirow{2}{*}{\textbf{Metric}} & \multicolumn{4}{c|}{\textbf{Chem42}} & \multicolumn{3}{c|}{\textbf{Baseline}} \\
         
         &  &  & 190M & 387M & 597M & 1002M & 140M & 184M & 507M\\
        
        \hline
        Pharmacokinetic & BBBP & ROC-AUC $\uparrow$ & 0.770 & \textbf{0.775} & 0.763 & 0.762 & 0.769 & 0.778 & 0.76\\
        
        \hline
        \multirow{4}{*}{Bioactivity} & BACE & ROC-AUC $\uparrow$ & 0.848 & 0.851 & 0.853 & \textbf{0.867} & 0.861 & - & 0.852 \\
        & HIV & ROC-AUC $\uparrow$ & 0.774 & \textbf{0.789} & 0.774 & 0.732 & 0.776 & 0.788 & 0.797\\
        & MUV & ROC-AUC $\uparrow$ & 0.824 & 0.850 & \textbf{0.853} & 0.712 & 0.816 & 0.838 & 0.832\\
        & PCBA & PRC-AUC $\uparrow$ & \textbf{0.892} & - & - & 0.886 & 0.257 & 0.882 & 0.314\\
        
        \hline
        \multirow{3}{*}{Toxicity} & TOX21 & ROC-AUC $\uparrow$ & 0.836 & \textbf{0.873} & 0.862 & 0.870 & 0.836 & 0.854 & 0.859\\
        & SIDER & ROC-AUC $\uparrow$ & 0.693 & 0.698 & \textbf{0.716} & 0.578 & 0.668 & 0.681 & 0.686\\
        & CLINTOX & ROC-AUC $\uparrow$ & \textbf{0.976} & 0.962 & 0.939 & 0.971 & 0.960 & 0.953 & 0.949\\
        
        \hline
        \multirow{4}{*}{Physicochemical} & ESOL & RMSE $\downarrow$ & 0.480 & \textbf{0.468} & 0.511 & 0.470 & 0.543 & 0.447 & 0.514\\
        & FREESOLV & RMSE $\downarrow$ & 0.638 & \textbf{0.582} & 0.673 & 1.023 & 0.820 & 0.584 & 0.739\\
        & LIPO & RMSE $\downarrow$ & 0.568 & \textbf{0.506} & 0.519 & 0.578 & 0.586 & 0.534 & 0.579\\
        
        \hline
        Molecular Binding & PDBBind Full & RMSE $\downarrow$ & 0.743 & 0.704 & 0.697 & \textbf{0.674} & 0.723 & 0.705 & 0.706\\
        \hline
        
    \end{tabular}
    \caption{Comparing Chem42 and Baseline models on Moleculenet \cite{wu2018moleculenet} benchmark \textbf{Bold}: best performance.}
    \label{tab:extended_moleculenet_benchmark}
\end{table}

    \begin{table}[ht!]
    \centering
    \scriptsize
    \begin{tabular}{|c|c|c|c|c|c|c|c|c|c|c|c|c|c|}
        \hline
        \multirow{2}{*}{\textbf{Category}} & \multirow{2}{*}{\textbf{Dataset}} & \multirow{2}{*}{\textbf{Metric}} & \multicolumn{4}{c|}{\textbf{Chem42}} & \multicolumn{3}{c|}{\textbf{Baseline}}  \\
         
         &  &  & 190M & 387M & 597M & 1002M & 140M & 184M & 507M \\
        
        \hline
        \multirow{6}{*}{Absorption} & Caco2\_Wang & MAE $\downarrow$ & 0.296 & \textbf{0.286} & 0.289 & \textbf{0.286} & 0.295 & 0.291 & 0.321\\
        & Bioavailability\_Ma & ROC-AUC $\uparrow$ & 0.658 & 0.725 & \textbf{0.745} & 0.739 & 0.707 & 0.707 & 0.713\\
        & Lipophilicity\_AstraZeneca & MAE $\downarrow$ & 0.493 & \textbf{0.455} & 0.464 & 0.463 & 0.490 & 0.470 & 0.474 \\
        & Solubility\_AqSolDB & MAE $\downarrow$ & 0.717 & 0.712 & \textbf{0.706} & 0.735 & 0.746 & 0.722 & 0.732\\
        & HIA\_Hou & ROC-AUC $\uparrow$ & 0.993 & 0.991 & 0.995 & 0.995 & \textbf{0.996} & 0.994 & 0.990\\
        & Pgp\_Broccatelli & ROC-AUC $\uparrow$ & 0.910 & 0.927 & \textbf{0.930} & 0.855 & 0.904 & 0.909 & 0.913 \\
        
        \hline
        \multirow{3}{*}{Distribution} & BBB\_Martins & ROC-AUC $\uparrow$ & 0.919 & 0.932 & 0.932 & \textbf{0.995} & 0.919 & 0.940 & 0.921\\
        & PPBR\_AZ & MAE $\downarrow$ & 8.152 & \textbf{7.479} & 7.677 & 7.662 & 7.974 & 8.126 & 7.943\\
        & VDss\_Lombardo & Spearman $\uparrow$ & 0.702 & 0.711 & \textbf{0.739} & 0.694 & 0.719 & 0.691 & 0.696\\
        
        \hline
        \multirow{6}{*}{Metabolism} & CYP2C9\_Veith & PRC-AUC $\uparrow$ & 0.744 & \textbf{0.768} & 0.767 & 0.765 & 0.743 & 0.756 & 0.764\\
        & CYP2D6\_Veith & PRC-AUC $\uparrow$ & 0.667 & \textbf{0.684} & 0.678 & 0.680 & 0.626 & 0.668 & 0.669\\
        & CYP3A4\_Veith & PRC-AUC $\uparrow$ & \textbf{0.881} & 0.876 & \textbf{0.881} & 0.875 & 0.871 & 0.888 & \textbf{0.881}\\
        & CYP2C9\_Sub.\_Car.Man. & PRC-AUC $\uparrow$ & 0.514 & 0.501 & 0.509 & 0.548 & 0.454 & \textbf{0.557} & 0.457\\
        & CYP2D6\_Sub.\_Car.Man. & PRC-AUC $\uparrow$ & 0.718 & 0.702 & \textbf{0.740} & 0.522 & 0.630 & 0.724 & 0.716\\
        & CYP3A4\_Sub.\_Car.Man. & PRC-AUC $\uparrow$ & 0.620 & \textbf{0.656} & 0.637 & 0.595 & 0.567 & 0.648 & 0.611\\
        
        \hline
        \multirow{3}{*}{Excretion} & Half\_Life\_Obach & Spearman $\uparrow$ & 0.596 & 0.613 & 0.595 & 0.615 & 0.576 & \textbf{0.627} & 0.539\\
        & Clearance\_Hepatocyte\_AZ & Spearman $\uparrow$ & 0.424 & 0.459 & 0.490 & \textbf{0.499} & 0.427 & 0.498 & 0.484\\
        & Clearance\_Microsome\_AZ & Spearman $\uparrow$ & 0.620 & 0.614 & 0.622 & 0.633 & 0.611 & \textbf{0.640} & 0.597\\

        \hline
        \multirow{4}{*}{Toxicity} & LD50\_Zhu & MAE $\downarrow$ & 0.530 & \textbf{0.520} & 0.522 & 0.524 & 0.565 & 0.536 & 0.535\\
        & hERG & ROC-AUC $\uparrow$ & 0.859 & 0.847 & 0.815 & 0.830 & 0.819 & \textbf{0.888} & 0.822\\
        & AMES & ROC-AUC $\uparrow$ & 0.833 & \textbf{0.850} & \textbf{0.850} & 0.846 & 0.836 & 0.833 & 0.824\\
        & DILI & ROC-AUC $\uparrow$ & 0.929 & 0.887 & 0.896 & 0.923 & 0.874 & \textbf{0.947} & 0.887\\
        
        \hline
    \end{tabular}
    \caption{Comparing Chem42 and Baseline models on ADMET \cite{huang2021therapeutics} benchmark. \textbf{Bold}: best performance.}
    \label{tab:extended_admet_benchmark}
\end{table}

\begin{table}[ht!]
    \centering
    \scriptsize
    \begin{tabular}{|c|c|c|c|c|c|}
        \hline
        \# \textbf{Params} & \# \textbf{Layers} & \textbf{Hidden Size} & \textbf{Filter Size} & \# \textbf{Heads} & \# KV \textbf{Heads}  \\
        \hline
        38M & 24 & 384 & 1024 & 8 & 4 \\
        44M & 24 & 384 & 1280 & 12 & 4 \\
        53M & 24 & 448 & 1280 & 16 & 4 \\
        62M & 24 & 448 & 1568 & 16 & 4 \\
        72M & 24 & 512 & 1536 & 16 & 4 \\
        81M & 24 & 512 & 1792 & 16 & 4 \\
        133M & 24 & 704 & 2048 & 16 & 4 \\
        148M & 24 & 768 & 2048 & 16 & 4 \\
        184M & 24 & 768 & 2688 & 16 & 4 \\
        191M & 24 & 768 & 2816 & 16 & 4 \\
        229M & 24 & 896 & 2816 & 16 & 4 \\
        270M & 24 & 1024 & 2816 & 16 & 4 \\ 
        298M & 24 & 1024 & 3200 & 16 & 4 \\
        327M & 24 & 1024 & 3584 & 16 & 4 \\
        387M & 24 & 1088 & 4032 & 16 & 4 \\
        414M & 24 & 1152 & 4032 & 16 & 4 \\
        507M & 24 & 1280 & 4480 & 20 & 4 \\
        \hline
        
    \end{tabular}
    \caption{Model architecture of the models used during the scaling law experiments. We configure hyperparameters with \(\mu\)Transfer using best values obtained from hyperparameter sweep.}
    \label{tab:scaling_law_models}
\end{table}

%\begin{figure}[ht]
%    \centering
%    \begin{subfigure}[t]{0.95\textwidth}
%        \centering
%        \includegraphics[width=\textwidth]{figs/canonical_shuffled.pdf}
%        \caption{Atom Level}
%    \end{subfigure}
%    \hfill
%    \begin{subfigure}[t]{0.95\textwidth}
%        \centering
%        \includegraphics[width=\textwidth]{figs/deepchem_canonical_shuffled.pdf}
%        \caption{Motif Level}
%    \end{subfigure}
%    \caption{Most frequent 100 tokens for atom and motif level tokenizers.}
%    \label{fig:token_freq}
%\end{figure}

%\begin{figure*}
%    \centering
%    \includegraphics[width=1\columnwidth]{figs/tpp.pdf}
%    \caption{Scaling Law Experiments. We plot validation loss against token per parameter TPP.}
%    \label{fig:scaling_law}
%\end{figure*}

%\begin{figure*}
%    \centering
%    \includegraphics[width=1\columnwidth]{figs/mup.pdf}
%    \caption{$\mu$P Hyperparameter Sweep. We conduct 200 runs for hyperparameter sweep over standart deviation for parameter initializers, embedding multiplier, output layer logit multiplier and peak learning rate}
%    \label{fig:mup_sweep}
%\end{figure*}

\end{document}